\documentclass[letterpaper, 10pt, journal, twoside]{IEEEtran}  % Comment this line out if you need a4paper

\IEEEoverridecommandlockouts                              % This command is only needed if 
\usepackage[utf8]{inputenc}
\usepackage[T1]{fontenc}
\usepackage{cite}
\usepackage[linesnumbered, ruled]{algorithm2e}

\usepackage{graphics} % for pdf, bitmapped graphics files
\usepackage{graphicx}
\usepackage{float}
\usepackage{epsfig} % for postscript graphics files
\usepackage{times} % assumes new font selection scheme installed
\usepackage{amsmath} % assumes amsmath package installed
\usepackage{amssymb}  % assumes amsmath package installed
\usepackage{bm}

\usepackage{pgfplots}
\DeclareUnicodeCharacter{2212}{−}
\usepgfplotslibrary{groupplots,dateplot}
\usetikzlibrary{patterns,shapes.arrows}
\pgfplotsset{compat=newest}

\usepackage{multirow}
\usepackage[normalem]{ulem}
\usepackage{units}

\usepackage[printonlyused]{acronym}
\newacro{MDP}{Markov Decision Process}
\newacro{RL}{Reinforcement Learning}
\newacro{Meta-RL}{Meta Reinforcement Learning}
\newacro{PPO}{Proximal Policy Optimization}
\newacro{MPC}{Model Predictive Control}
\newacro{CMA-ES}{Covariance Matrix Adaptation Evolution Strategy}
\newacro{MLP}{Multi Layer Perceptron}
\newacro{MAML}{Model-Agnostic Meta-Learning}
\newacro{MCOT}{Mechanical Cost Of Transport}

\useunder{\uline}{\ul}{}

\title{
Meta Reinforcement Learning for \\Optimal Design of Legged Robots
}

\author{\'Alvaro Belmonte-Baeza$^*$, Joonho Lee$^*$, Giorgio Valsecchi and Marco Hutter% <-this % stops a space
\thanks{Manuscript received: June 6, 2022; Revised: September 6, 2022; Accepted: September 25, 2022.}
\thanks{This paper was recommended for publication by Editor Jens Kober upon evaluation of the Associate Editor and Reviewers' comments.}
\thanks{The project leading to these results has received funding from "la Caixa" Foundation (ID 100010434), under agreement LCF/BQ/EU20/11810067, ESA Contract Number 4000131516/20/NL/MH/ic, and from the ERC No 852044.}% <-this % stops a space

\thanks{*\'Alvaro Belmonte-Baeza and Joonho Lee contributed equally. \textit{(Corresponding authors: \'Alvaro Belmonte-Baeza; Joonho Lee)}. All authors are with the Robotic Systems Lab (RSL), ETH Zürich, Switzerland.
}
\thanks{Digital Object Identifier (DOI): see top of this page.}
}

\begin{document}
\maketitle
%\thispagestyle{empty}
%\pagestyle{empty}

%%%%%%%%%%%%%%%%%%%%%%%%%%%%%%%%%%%%%%%%%%%%%%%%%%%%%%%%%%%%%%%%%%%%%%%%%%%%%%%%
\begin{abstract}

The process of robot design is a complex task and the majority of design decisions are still based on human intuition or tedious manual tuning. A more informed way of facing this task is computational design methods where design parameters are concurrently optimized with corresponding controllers. Existing approaches, however, are strongly influenced by predefined control rules or motion templates and cannot provide end-to-end solutions. In this paper, we present a design optimization framework using model-free meta reinforcement learning, and its application to the optimizing kinematics and actuator parameters of quadrupedal robots. We use meta reinforcement learning to train a locomotion policy that can quickly adapt to different designs. This policy is used to evaluate each design instance during the design optimization. We demonstrate that the policy can control robots of different designs to track random velocity commands over various rough terrains. With controlled experiments, we show that the meta policy achieves close-to-optimal performance for each design instance after adaptation. Lastly, we compare our results against a model-based baseline and show that our approach allows higher performance while not being constrained by predefined motions or gait patterns.
% In addition, benchmarking against previous design optimization approaches, we show that by introducing the flexibility and robustness of learning-based control, we can perform design optimization unconstrained to predefined tasks and motions.
\end{abstract}


%%%%%%%%%%%%%% Article sections. Included from separate files %%%%%%%%%%%%%%%%%%

\section{INTRODUCTION}
\label{sec:introduction}

\IEEEPARstart{D}{esigning} a robot is an arduous task, since there are many parameters that affect its final performance. In the case of legged robots, these design parameters can include limb lengths, drive-train parameters such as gear ratio, and control parameters such as gait parameters and duration~\cite{chadwick2020Vitruvio}. The wide range of continuous and discrete design parameters results in a combinatorial problem with often unclear correlations between the design parameters and the resulting robot behavior.

Unfortunately, literature on the design principles of the legged robot is very sparse.
In order to make design decisions, designers often rely on approximations, simulations, or bio-inspired solutions~\cite{MITCheetah2012}. Some examples of quadrupedal robots designed in this conventional paradigm are Mini Cheetah \cite{katz2019mini}, HyQ \cite{Semini2011HyQ}, or ANYmal \cite{Hutter2017ANYmal}.
Some mention that the range of motion and inertia are considered in leg design~\cite{katz2019mini} or certain performance goals~\cite{Semini2011HyQ} are examined, but often it is unclear how the final values are determined.

For a more quantitative approach to robot design, computational optimization methods have been introduced to search for an optimal design~\cite{papalambros_wilde_2000}.
In this paradigm, designing a robot is formulated as a bilevel optimization problem. Optimizing the design parameters is the outer optimization problem, and determining optimal control parameters for each design instance is the inner problem. Usually, the inner loop entails multiple sub-objectives, and is generally not differentiable with respect to the design parameters.

\begin{figure}
    \centering
    \includegraphics[width=\columnwidth]{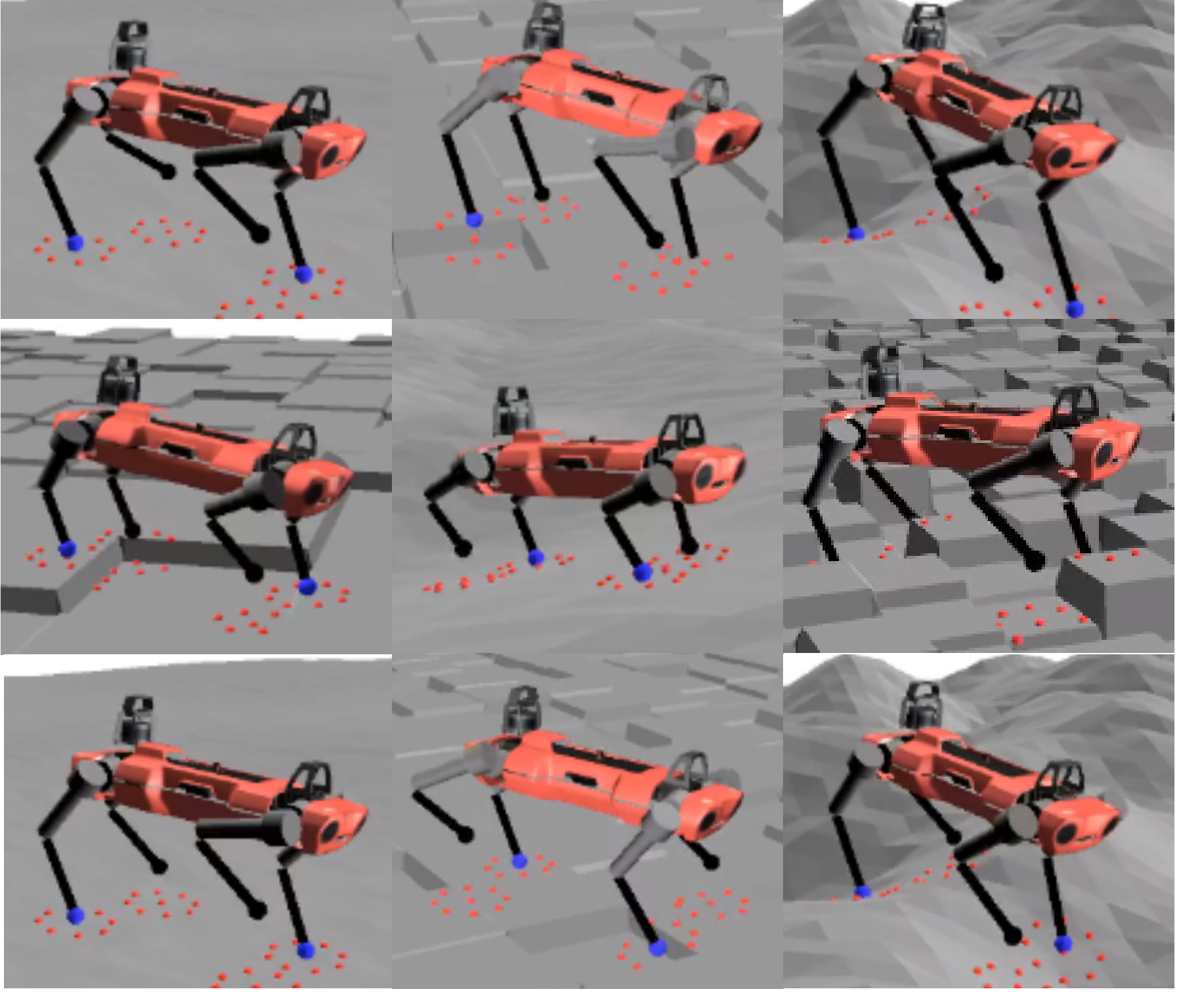}
    \caption{Optimal designs for different situations.}
    \label{fig:intro}
\end{figure}

Existing works can be broadly separated into two categories: Gradient-based and gradient-free methods.
Gradient-based approaches try to define a differentiable relationship between the control performance (the result of the inner loop) and the design parameters.
In Ha et al.~\cite{Ha2017JointOptimization}, the relationship between motion parameters and design parameters is defined via the implicit function theorem such that the gradient computation is feasible.
De Vincenti et al.~\cite{devincenti2021control} directly differentiated inverse dynamics-based whole body controllers to optimize the leg design of a quadrupedal robot.
Dinev et al.~\cite{dinev2021co} also employed gradient-based optimization to optimize the base and leg shape of a quadrupedal robot.

Gradient-free methods are better suited for non-convex problems. Starting from the computer graphics community~\cite{Sims1994Evolve}, this paradigm has extended its scope to mechanical design optimization. In~\cite{digumarti2014concurrent}, joint optimization of design and control parameters is performed to maximize the linear speed of a quadrupedal robot using \ac{CMA-ES}~\cite{hansen2001completely}. It was also employed by Ha et al.~\cite{Ha2016TaskBasedDO} to optimize a 2D model of a legged robot. Chadwick et al.~\cite{chadwick2020Vitruvio} introduced a framework for obtaining optimal leg designs for walking robots using a genetic algorithm. 

All the aforementioned works have one common building block: they rely on model-based control approaches. The model-based control methods are generalizable and intuitive, but they have several limitations for complex systems like legged robots. 
Firstly, they often rely on simplified models to reduce complexity. For example, the trajectory optimization method used in \cite{chadwick2020Vitruvio} relies on centroidal dynamics and ignores limb masses. Thus, the design parameters are optimized under the significant dynamics mismatch and the resulting controller cannot be used on the physical system.
Secondly, the resulting motions rely on handcrafted primitives and are restricted to predetermined tasks and trajectories, e.g., predefined gait patterns or base trajectory~\cite{chadwick2020Vitruvio, dinev2021co, devincenti2021control}. 
Lastly, since the motion parameters and simplified dynamics models are often developed/tuned for a certain instance by hand, it is hard to claim that the optimized motion is truly optimal for each design.

An alternative control method that has recently gained a lot of attention for robot control is Model-Free \ac{RL}~\cite{sutton2018reinforcement}. The field of legged locomotion has been especially active, and has shown very promising results. 

Lee et al.~\cite{Lee2020Challenging} demonstrated that it is possible to learn a control policy for blind quadrupedal locomotion over challenging, natural environments, and Miki et al.~\cite{miki2022learning} extended this to also leverage exteroceptive perception of the environment, resulting in an increase in robustness and speed. These last contributions validated the viability of the \ac{RL} in this context. 

\ac{RL} can be an ideal solution to solve the inner optimization problem of the design optimization since we can obtain a control policy without model simplification and heuristics. However, \ac{RL} has been barely used in the design optimization literature, with examples in simpler 2D scenarios~\cite{Ha2018designrl}, or multi-object airfoil optimization for better aerodynamic performance~\cite{Hui2021RLAeroDesign}.

We extend the work carried out by Won and Lee~\cite{Won2019MorphCon}, where they handle changes in body size and proportions of virtual characters. A single controller is trained to control characters with different dimensions on the run, without re-training. This method suggests that it is possible to obtain a control policy capable of managing a range of designs. 

Although a good starting point, a naive multi-task \ac{RL}, where a policy is trained with randomly sampled tasks, often results in a policy that performs "generally well" in the task space and cannot reach the performance of a specialized policy for a certain task~\cite{parisotto2015actor}. 
Using a "generally performant" controller won't suffice for the design optimization because each design parameter needs to be evaluated with the best performance.
Hence, we want to train a policy that achieves the performance of a specialized policy with the least amount of fine-tuning effort.

Meta-learning~\cite{schmidhuber1987evolutionary} has proven to be a promising direction to allow quick adaptation of a neural network model to a certain task by leveraging information from other similar tasks~\cite{Finn2017MAML},~\cite{gupta2018meta}. This is done by training a single model in a range of different tasks, shaping the model parameter space in a way that favors few-shot adaptation to newly encountered tasks during test time.
Finn et al.~\cite{Finn2017MAML} demonstrated that a \ac{Meta-RL} training enables a fast-adapting control policy.

Based on these insights, we present a framework for design optimization that evaluates each design parameter using a fast-adapting \ac{Meta-RL} policy.
We demonstrate its application to the optimization of quadrupedal robots. Fig.~\ref{fig:intro} shows designs optimized for different objective-environment pairs.

Our main contributions are: \begin{itemize}
    \item A design optimization framework using a robust, adaptive neural network controller.
    \item A \ac{Meta-RL} approach to obtain a locomotion control policy that can be easily fine-tuned for different robot designs.
    \item Experimental results in simulation demonstrating the influence of different design objectives and operating environments on the robot's design.
\end{itemize}

\begin{figure*}[t]
    \centering
    \includegraphics[width=0.8\textwidth]{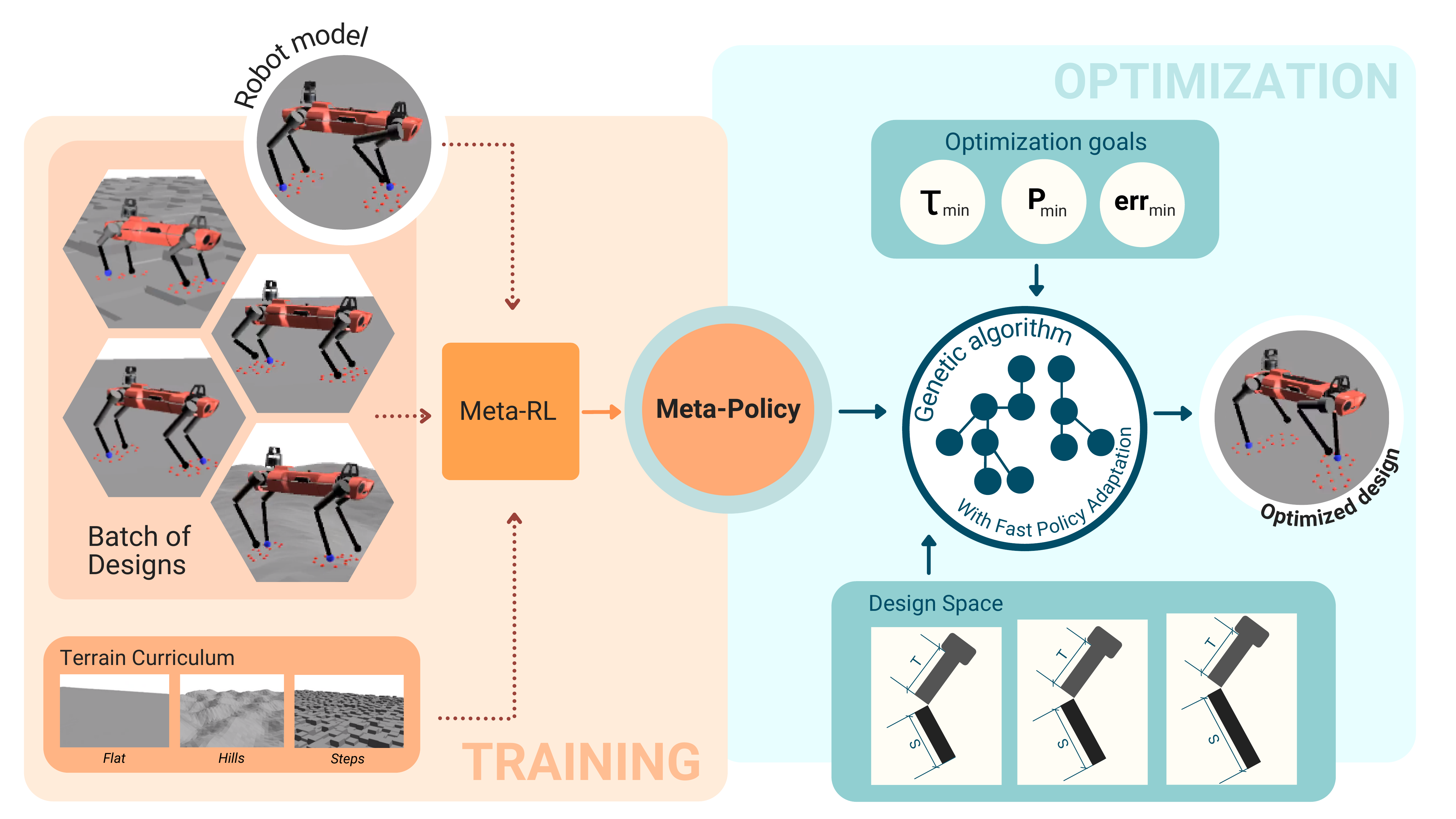}
    \caption{Overview of the proposed design optimization pipeline. We first train a control policy that adapts to different design instances within a range of design parameters via \ac{Meta-RL}, then perform the design optimization using a genetic algorithm}
    \label{fig:overview}
\end{figure*}

\section{METHOD}
\label{sec:method}

Our approach aims to exploit the robustness and versatility of \ac{RL}-based control methods to obtain a locomotion policy that is not constrained to a specific motion and environment. This allows us to perform design optimization based on data obtained from a wide variety of motions, resulting in an optimized design not constrained to a specific situation.

Fig.~\ref{fig:overview} shows an overview of our approach. Our method consists of two different phases: First, we use \ac{Meta-RL} to train a policy with randomly sampled design parameters and terrains. We used the \ac{RL} environment by Lee et al.~\cite{Lee2020Challenging} with terrain curriculum. Secondly, the trained policy is used to evaluate different designs and find a design that maximizes a design objective. Importantly, since the policy training and the design optimization are separated, the trained meta-policy can be reused for different design optimization tasks.

In this section, we describe our \ac{Meta-RL} approach to train an adaptive locomotion policy and the implementation of our design optimization framework.

%%%%%%%%%%%%%%%%%%%%%%%%%%%%%%%%%%%%%%%%%%%%%%%%%%%%%%%%%%%%%%%%%%%%%%%%%%%%%%%%%%%%%
\subsection{Markov Decision Process}
\label{sec:method:rlenv}
We model the locomotion control problem as a \ac{MDP}. \ac{MDP} is a mathematical framework for formulating a discrete-time decision-making process which is commonly used in \ac{RL}.
An \ac{MDP} is defined by a tuple of state space $\mathcal{S}$, action space $\mathcal{A}$, the transition probability density $\mathcal{P}(s_{t+1}|s_t,a_t)$, and a reward function $\mathcal{R}(s_t,a_t, s_{t+1}): \mathcal{S}\times\mathcal{A}\times\mathcal{S}\to\mathbb{R}$. Every timestep, the learning agent receives a state $s_t \in \mathcal{S}$ from the environment and takes an action $a_t \in \mathcal{A}$ depending on its policy $\pi_{\theta}(a_t|s_t)$, receiving a reward value $r_t$. 

The objective of \ac{RL} is to obtain an optimal policy $\pi_{*}$ that maximizes the cumulative discounted rewards $\mathbb{E}[\sum_{t=j}^{\infty}\gamma^t r_t]$ throughout interactions with its environment in an iterative fashion, where $\gamma$ is known as the discount factor.

We define the \ac{MDP} akin to the "teacher policy training" from \cite{Lee2020Challenging} and \cite{miki2022learning}. $s_t$ is composed of the velocity command, linear and angular body velocity, joint states,  frequency and phase of the gait pattern generators for each foot (the policy learns to modulate periodic leg phases), and the two last actions taken by the policy. Since our design optimization is conducted in simulation, we also make use of the privileged observations consisting of contact states from the different parts of the robot, and terrain height information around each foot.Additionally, our observation vector includes the design parameters, as in \cite{Won2019MorphCon}.

Our reward function consists of velocity tracking rewards for the command $\langle v_x^{\text{target}}, v_y^{\text{target}}, \omega_z^{\text{target}} \rangle$ like in \cite{miki2022learning}, which denotes linear velocity in $x$, $y$ direction and angular velocity along $z$-axis in base frame. We also favor stability of the base by penalizing velocities orthogonal to the command. We encourage stepping by rewarding the number of feet not in contact with the ground, and penalize collisions with the rest of the parts of the robot. To encourage smooth motion, we penalize differences both in outputs and velocity between time steps. Finally, we also penalize joint torques to increase efficiency and prevent damage. A complete definition of the reward function is provided in Table \ref{table:reward} in the appendix.

The policy $\pi_{\theta}(a_t|s_t)$ is modeled as a Gaussian policy, i.e., $a_t \sim \mathcal{N}(m_\theta(s_t), \sigma_\theta)$ using a \ac{MLP} for $m_\theta(\cdot)$ and a state-independent $\sigma_\theta$.
We use \ac{PPO} \cite{ppo} for policy optimization.
%
%%%%%%%%%%%%%%%%%%%%%%%%%%%%%%%%%%%%%%%%%%%%%%%%%%%%%%%%%%%%%%%%%%%%%%%%%%%%%%%%%%%%%%%
\subsection{Fast Adaptation with Meta-learning}
\label{sec:method:meta}
The \ac{MAML}~\cite{Finn2017MAML} approach is used to train our meta-policies. \footnote{We followed the author's implementation: github.com/cbfinn/maml}
In this framework, we define a distribution of tasks $p(\mathcal{T})$ that we want the policy to be able to adapt to. $p(\mathcal{T})$ is a uniform distribution over the design parameters in our setup.
Algorithm \ref{algo:meta-training} describes the training process.
Each policy update involves training the policy for $M$ different tasks sampled separately from $p(\mathcal{T})$ (Lines 4-9), gathering samples with fine-tuned policies $\pi_{\theta_i'}$ for each ($\mathcal{T}_i$), and updating the policy using the aggregated data (Line 10).
This approach enables fast fine-tuning to a task from $p(\mathcal{T})$ with a small amount of data possible during test time. 

%%%%%%%%%%%%%%%%%%%%%%%%%%%%%%%%%%%%%%%%%%%%%%%%%%%%%%%%%%%%%%%%%%%%%%%%%%%%%%%%%%%%%%%%
\subsection{Design Optimization}

The goal of the design optimization is obtaining a set of design parameters that maximizes a given fitness function $f(\mathcal{T}) \in \mathbb{R}$.  Due to the non-differentiable way we evaluate each design parameter, our design optimizer should be a gradient-free algorithm.
We use \ac{CMA-ES}~\cite{hansen2001completely} for the optimization, which has been widely used in the design optimization literature as stated in section \ref{sec:introduction}.

The fitness function is the Monte-Carlo estimation of a performance metric $\mathcal{C}(s_t):\mathcal{S}\to\mathbb{R}$, i.e., 
\begin{equation*}
    f(\mathcal{T}) = \mathbb{E}_{s_t \sim \xi ( \pi_{\theta_\mathcal{T}} ) } [ - \mathcal{C}(s_t)]
\end{equation*}
, where $\xi({\pi(\mathcal{T})})$ denotes trajectories generated by a policy $\pi_\mathcal{T}$ fine-tuned for $\mathcal{T} \sim p(\mathcal{T})$.
We will define $\mathcal{C}$ in the next section.

For every $\mathcal{T}$, we perform an adaptation of our meta-policy in the optimization loop, in order to achieve optimal performance for each design. The inclusion of these adaptation steps is described in Algorithm \ref{algo:optimization}.

\begin{algorithm}[t] \label{algo:meta-training}
\SetAlgoLined
 \SetKwInOut{Input}{Input}

  \Input{Parametrized policy $\pi_{\theta}$, Distribution over tasks $p(\mathcal{T})$, Number of policy updates $N$, Meta-batch size $M$, length of collected rollouts $K$. Step-size hyperparameters $\alpha, \beta$}
  Initialize $\theta$\;
  \For{N policy updates}{
     Sample batch of $M$ design parameter tuples $\mathcal{T}_i \sim p(\mathcal{T})$\;
     \ForEach{$\mathcal{T}_i$}{
        Sample policy rollouts of length K $\mathcal{D}  = \{(s_1,a_1,r_1,s_2, \dots, s_K) \}$\;
        Compute adapted parameters for current task:\\
        $\theta_i^{\prime} = \theta - \alpha \nabla_{\theta}\mathcal{L}_{\mathcal{T}_i}(\pi_{\theta})$\;
        Sample new trajectories $\mathcal{D}^{\prime}_i$ using adapted policy $\pi_{\theta^{\prime}_i}$ in $\mathcal{T}_i$\;
     }
     Update $\theta \leftarrow \theta - \beta \nabla_{\theta}\sum_{\mathcal{T_i}}\mathcal{L}_{\mathcal{T}_i}(\pi_{\theta^{\prime}_i})$, using the collected $\mathcal{D}^{\prime}_i$\;
     
  }
  \caption{Policy meta-training with MAML}
\end{algorithm}

\begin{algorithm}[t] \label{algo:optimization}
  \SetAlgoLined
  \SetKwInOut{Input}{Input}
  \SetKwData{NumGradUpdates}{num\_grad\_updates}
  
  \Input{Trained meta-policy $\pi_{\theta_0}$, Number of generations G, Initial design population $\mathcal{P}_{0}$, step-size hyperparameter $\alpha$, number of gradient updates $U$, lenght of collected rollouts $T$.}
  
  \For{k in [1...G]}{
      \ForEach{$p_i \in \mathcal{P}_{k}$}{
        Set current design to $p_i$\;
        Set policy parameters to initial value: $\theta \leftarrow \theta_0$\;
        \For{U gradient updates}{
            Sample policy rollouts of length T $\mathcal{D}  = \{(s_1,a_1,r_1,s_2, \dots, s_T) \}$\;
            Perform adaptation step:\\
            $\theta \leftarrow \theta - \alpha \nabla_{\theta}\mathcal{L}_{p_i}(\pi_{\theta})$\;
            }
         Compute fitness score for $p_i$ and store it\;
        }
        Update $\mathcal{P}$ using the computed scores. 
    }
  \caption{Design optimization with meta-policy}
\end{algorithm}
\section{EXPERIMENTAL RESULTS}
\label{sec:results}
In this section, we first validate the effectiveness of our \ac{Meta-RL} approach and then present the outcomes of the design optimization under different objectives and environments. The policy behavior and the optimized designs in motion can be seen in the supplementary video.

\subsection{Experimental Setup}\label{setup}

\subsubsection{Design Parameters}
As shown in Fig.~\ref{fig:overview}, Our goal is to optimize the design of a robot's legs. A base design comes from a robot currently being developed, which consists of ANYmal C \cite{anybotics_2021} main body with longer legs. The nominal length is \unit[350]{mm} for both thigh and shank links, on the basis of simplified considerations similar to those mentioned in \cite{MITCheetah2012}. This value is a starting point for the optimization algorithm and does not have to be accurate.

The parametrization of the design consists of $l_t, l_s \in [0.6, 1.4]$, which are scale factors for the nominal link lengths. For example, the shank length is $l_s \times \unit[350]{mm}$. $\langle l_t, l_s \rangle$ defines a task $\mathcal{T}$. We adapt the link masses by applying the same scale factor, since a cylinder's mass scales linearly with its length. 

Other parameters have also been considered, such as the gear ratio of the actuators, the geometry of the linkage transmission, the attachment point of the legs to the base, and the orientation of the first actuator. We decided to restrict the optimization to the link lengths for the more in-depth experiments, to make the results more understandable and to compare with previous works. In section \ref{sec:results:gear}, we will extend the parameter space to include the gear ratios of the actuators.

\subsubsection{Design Optimization Objectives}
We perform the design optimization as described in Algorithm \ref{algo:optimization} using different cost functions ($\mathcal{C}(\mathcal{T}) : \mathcal{T} \to \mathbb{R}$) to verify our framework. We define three optimization objectives to be minimized:

\begin{itemize}
    \item Velocity tracking: 
    \begin{equation*}
    \mathcal{C}_v = \sum_{t\in[0, T]} ((e_v)_t^2 + (e_\omega)_t^2 ),
    \end{equation*}
    \item Weighted joint torque:
        \begin{equation*}
 \mathcal{C}_{\tau} =\sum_{t\in[0, T]} w_t \left( \sum_{i\in\{1,.. 12\}} (\tau_i)_{t}^2 \right )
    \end{equation*}
    \item Weighted joint positive mechanical power:
    \begin{equation*}
\mathcal{C}_{p} = \sum_{t\in[0, T]} w_t \left( \sum_{i\in\{1,.. 12\}} \max((\dot{\phi}_i \tau_i)_{t}, 0.0) \right )
    \end{equation*}
\end{itemize}

, where $ e_v = \lVert v_{x,y}^{\text{target}} - v_{x,y} \rVert_2 $, $e_\omega = \lVert \omega_z^{\text{target}}  - \omega_{z} \rVert_2$, and $w_t = \exp( 1.5  ((e_v)_t^2 +(e_\omega)_t^2))$. $\tau$ and $\dot{\phi}$ denote joint torque and velocity, respectively. We clip $w_t$ to a maximum value of 100 for stability of the optimization process.

When defining the optimality of a robot design, there is an important trade-off to consider: performance versus efficiency.
Strong and versatile robots usually consume more power or require higher joint torques.
With the weighting factor $w_t$, we account for this trade-off. When the tracking error ($e_v, e_\omega$) is large, $w_t$ grows exponentially.
This weighting leads to a Pareto optimum between minimizing the selected metric, while still maintaining a good performance level.

The trade-off is often not considered in existing model-based approaches like \cite{chadwick2020Vitruvio} since they rely on pre-defined kinematic trajectories. They are inherently bound by the motion generator that does not consider different designs.
On the other hand, we can explicitly optimize for the trade-off since our framework can generate optimal motions for any design instances from $p(\mathcal{T})$.

\subsubsection{Implementation Details}
The training environments are implemented using Raisim \cite{raisim} simulator. In addition, a simplified model for the velocity and torque limits of the real actuator is included in the simulation.
All the policies are trained for $N$=2000 epochs using the same hyperparameters for \ac{PPO}, which are detailed in Table \ref{table:ppo} in the appendix. Each epoch runs 1000 training environments with random velocity commands and terrain parameters.

The design optimization is implemented using PyCMA~\cite{hansen2019pycma} library.
The design optimization is performed for $G$=30 generations, with a population of 35 different designs. For each member of the population, the meta-policy is adapted using $U$=5 \ac{PPO} updates with rollouts of length $T$=50. Then, 250 transitions are collected with the adapted policy to compute the fitness score for \ac{CMA-ES}.
The adaptation and score computations for each member of the population are done with data from 300 parallel simulated environments, each of them with a randomly sampled command and differently generated terrains.

The design optimization takes about 1.4 hours of wall clock time using a desktop machine (CPU: AMD Ryzen 7 4800h, GPU: Nvidia GeForce GTX 1650Ti, 16 GB Memory) without parallelization of the evaluation for each member of the population in the CMA-ES algorithm.

\subsection{Effect of Meta-learning on the Policy Adaptation}
\label{sec:results:meta}
We validate our \ac{Meta-RL} approach for training design-conditioned policies. We compare a policy trained as described in Algorithm~\ref{algo:meta-training} (\textbf{meta-policy}) against a naive policy trained over uniformly sampled design parameters (\textbf{naive-policy}).   

\begin{figure}[t]
    \centering
    \includegraphics[width=\columnwidth]{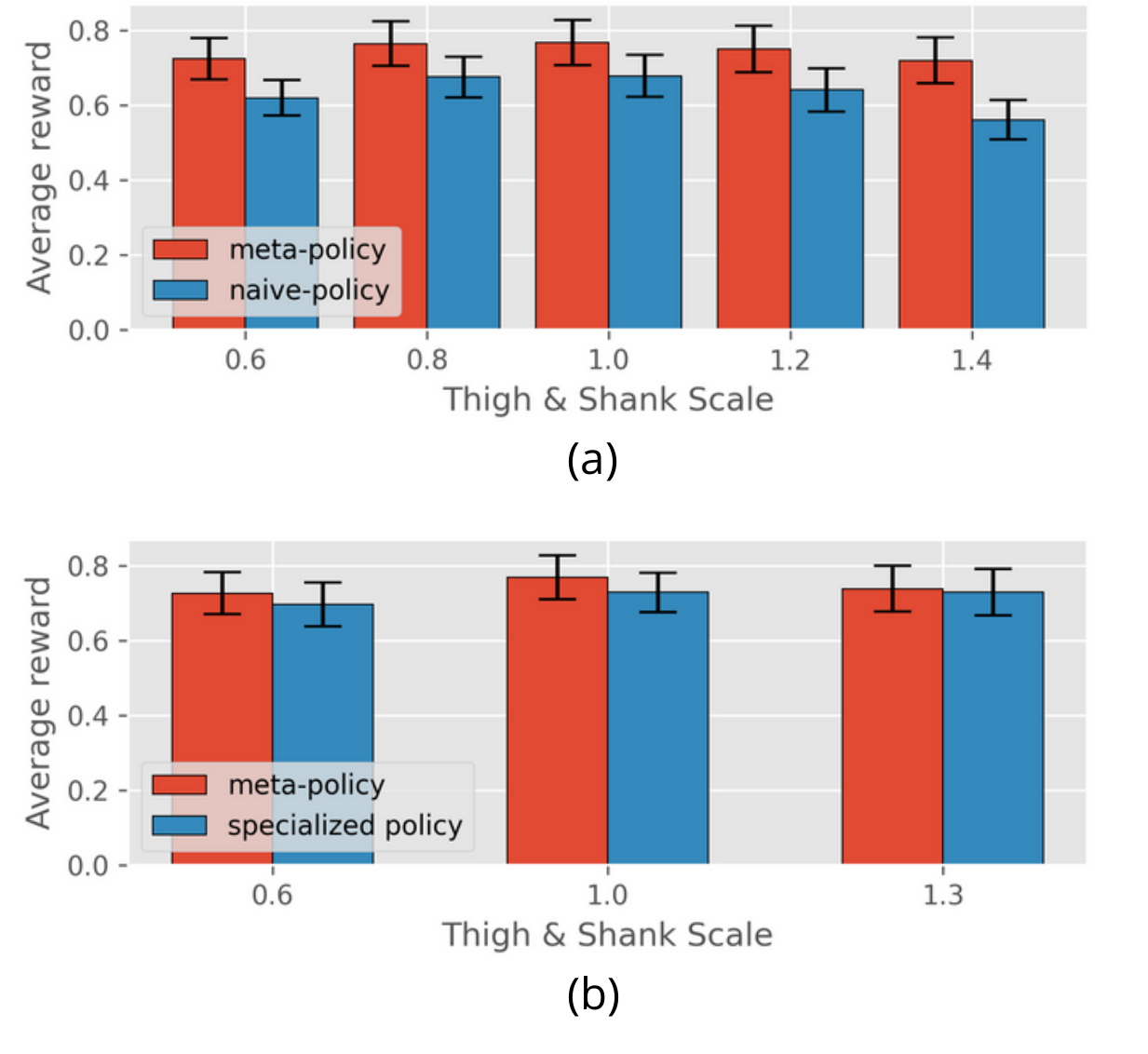}
    \caption{(\textbf{a}) Evaluation of average reward obtained by our meta-policy against a parametrized policy trained by randomizing the design parameters. (\textbf{b}) Evaluation of average reward obtained by our meta-policy against policies trained specifically for given designs.}
    \label{fig:meta}
\end{figure}

Fig. \ref{fig:meta}-(a) shows a comparison of the average reward obtained by the two policies across different parameters. The rewards are computed from 3000 rollouts of 500 time steps each. 
The meta-policy consistently outperforms the naive multi-task policy in all cases.

We further verify the performance of our meta-policy by comparing it against a set of policies trained for specific designs (\textbf{specialized policy}). The result is given in Fig. \ref{fig:meta}-(b). After the adaptation steps, the meta-policy reaches rewards comparable to the specialized policies, achieving close-to-optimal capabilities.

Based on this analysis, we use our meta-policy for the following design optimization experiments. For each evaluated design instance, we fine-tune our meta-policy.

\subsection{Design Optimization Using Different Objectives}
\label{sec:results:optimization-overview}

\begin{figure}
    \centering
    \includegraphics[width=\columnwidth]{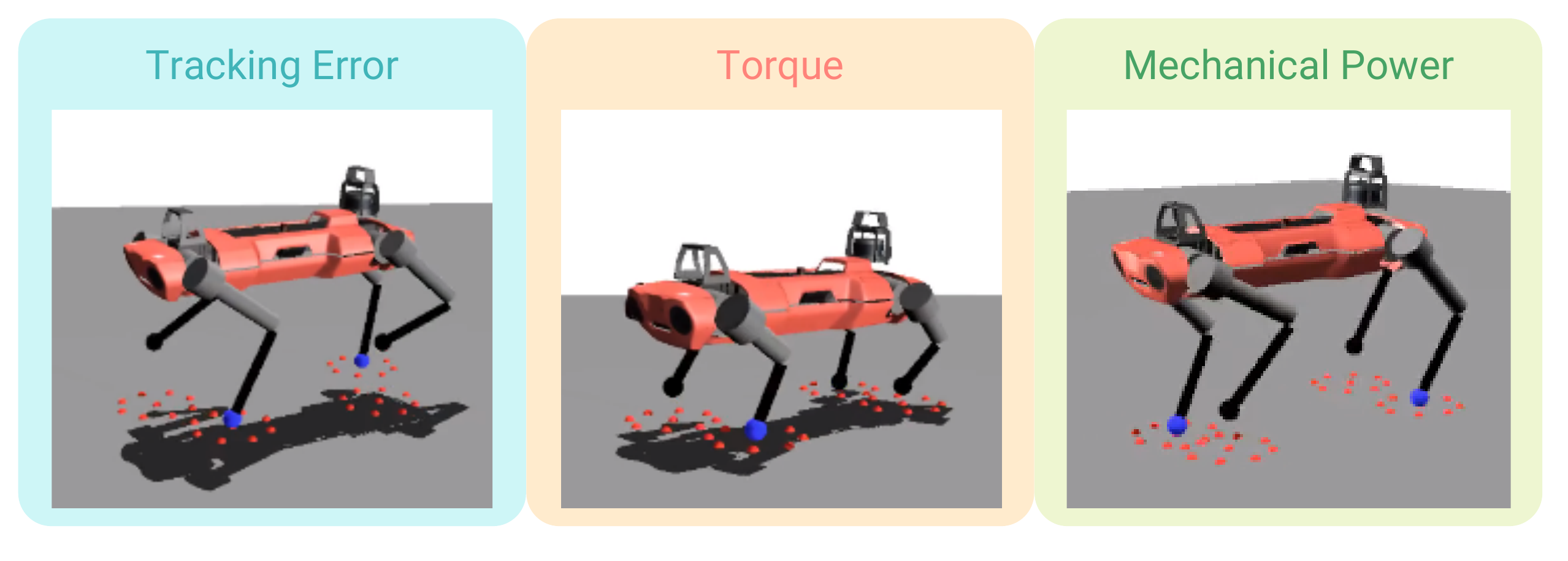}
    \caption{Optimal designs for different objectives on flat terrain.}
    \label{fig:optimal_flat}
\end{figure}

We conduct design optimization comparing three scenarios: (1) Performance-only, (2) Reducing joint torques, and (3) Reducing power consumption. $\mathcal{C}_v$, $\mathcal{C}_{\tau}$, and $\mathcal{C}_p$ are used, respectively.
The result is shown in Table~\ref{table:results}, Table~\ref{table:improvement}, and Fig. \ref{fig:optimal_flat}.

On flat terrain (Table~\ref{table:results}-Flat), the result for (1) is similar to the nominal design, which is designed by an engineer, with a slight increase in the thigh length (\unit[2]{\%}), so no major improvement is obtained. The torque-minimized design ($\mathcal{C}_\tau$) opts for the minimal possible leg length, thereby reducing the moment arm and total mass and inertia. This results in a drastic improvement of \unit[43.5]{\%} in the optimization score compared to the nominal design (Table~\ref{table:improvement}-Flat). 
% Insert meta vs ppo policy evaluation plot

\begin{table*}[ht]
\centering
\caption{Optimized link scales with respect to the nominal design}
\begin{tabular}{c|cccccccccccccc}
\hline
\textbf{Objective} & \multicolumn{2}{c}{\textbf{Flat}} & \multicolumn{2}{c}{\textbf{Easy Hills}} & \multicolumn{2}{c}{\textbf{Mid Hills}} & \multicolumn{2}{c}{\textbf{Hard Hills}} & \multicolumn{2}{c}{\textbf{Easy Steps}} & \multicolumn{2}{c}{\textbf{Mid Steps}} & \multicolumn{2}{c}{\textbf{Hard Steps}} \\ \hline
\textbf{}          & {\ul Thigh}     & {\ul Shank}     & {\ul Thigh}        & {\ul Shank}        & {\ul Thigh}        & {\ul Shank}       & {\ul Thigh}        & {\ul Shank}        & {\ul Thigh}        & {\ul Shank}        & {\ul Thigh}        & {\ul Shank}       & {\ul Thigh}        & {\ul Shank}        \\
\textbf{$\bm{\mathcal{C}_v}$}     & 1.02            & 0.99            & 1.01               & 1.01               & 1.06               & 1.03              & 1.23               & 1.18               & 1.05               & 1.0                & 1.07               & 1.06              & 1.21               & 1.17               \\
\textbf{$\bm{\mathcal{C}_{\tau}}$}    & 0.61            & 0.63            & 0.63               & 0.67               & 0.64               & 0.68              & 0.75               & 0.80               & 0.70               & 0.68               & 0.76               & 0.77              & 0.94               & 0.97               \\
\textbf{$\bm{\mathcal{C}_p}$}     & 1.05            & 0.94            & 1.07               & 0.95               & 1.06               & 0.93              & 1.10               & 0.97               & 1.04               & 0.93               & 1.07               & 0.96              & 1.17               & 1.13              
\end{tabular}
\label{table:results}
\end{table*}
\begin{table}[t]
\centering
\caption{Mean improvement in optimization objectives compared to the nominal design.}
\begin{tabular}{c|c|c|c|}
\cline{2-4}
                                          & $\bm{\mathcal{C}_v}$ & $\bm{\mathcal{C}_{\tau}}$ & $\bm{\mathcal{C}_p}$ \\ \hline
\multicolumn{1}{|c|}{\textbf{Flat}}       & 1.27\%              & 43.53\%              & 4.30\%              \\ \hline
\multicolumn{1}{|c|}{\textbf{Easy Hills}} & 2.16\%              & 43.85\%              & 5.07\%             \\ \hline
\multicolumn{1}{|c|}{\textbf{Mid Hills}}  & 4.32\%              & 39.72\%              & 3.01\%             \\ \hline
\multicolumn{1}{|c|}{\textbf{Hard Hills}} & 27.85\%             & 16.36\%              & 13.47\%            \\ \hline
\multicolumn{1}{|c|}{\textbf{Easy Steps}} & 4.50\%              & 37.47\%              & 4.10\%             \\ \hline
\multicolumn{1}{|c|}{\textbf{Mid Steps}}  & 6.45\%              & 28.98\%              & 5.47\%            \\ \hline
\multicolumn{1}{|c|}{\textbf{Hard Steps}} & 24.79\%             & 4.13\%              & 16.01\%            \\ \hline
\end{tabular}
\label{table:improvement}
\end{table}

The positive mechanical power ($\mathcal{C}_p$) minimization results in a design with similar legs, and with a higher thigh/shank proportion (1.05:0.94). This design aims to find a balance between reducing leg lengths to reduce the joint torque as in the previous case, but also avoiding high joint speed to limit power consumption. In addition, the longer thigh seeks for higher end-effector velocities with smaller joint speed. With this, we can get a \unit[4.3]{\%} reduction in weighted mechanical power consumption compared to the nominal design (Table~\ref{table:improvement}-Flat).

\subsection{Optimal designs for rough terrains}

\begin{figure}
    \centering
    \includegraphics[width=\columnwidth]{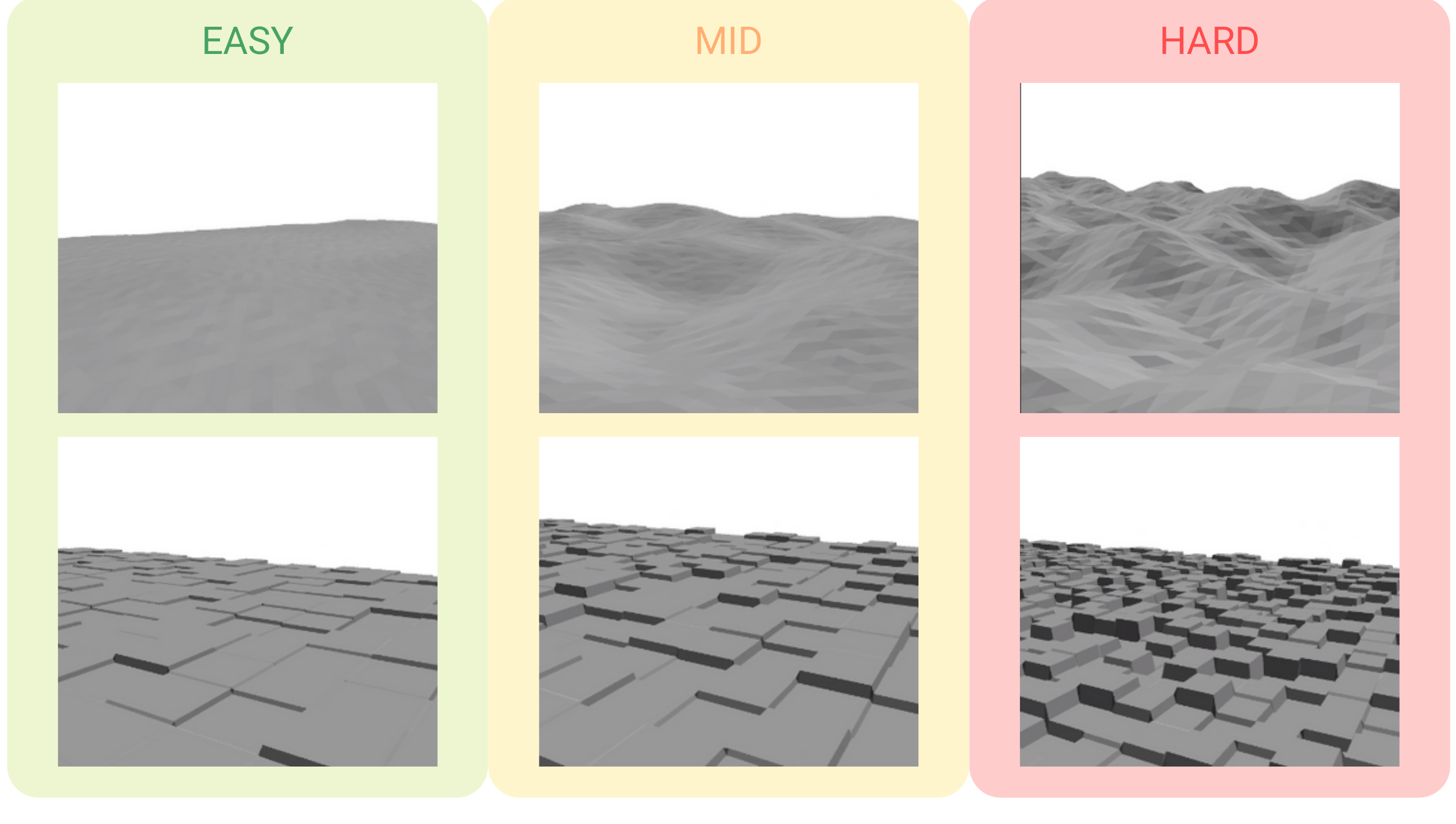}
    \caption{Examples of rough terrains with different difficulty levels used for the optimization.}
    \label{fig:terrains}
\end{figure}

We evaluate how different terrains affect the optimal leg design. 
We use parameterized terrains presented by~\cite{Lee2020Challenging}. 
Two types of terrains are simulated: hilly terrain and discrete steps. The former entails smooth transitions between slopes and flat terrains, and the latter simulates discrete height changes and foot-trapping while walking. The terrain parameters are randomized during training as well as friction coefficients for each foot. Examples of the terrains are shown in Fig. \ref{fig:terrains}. We modulate the difficulty level by changing the roughness, frequency, and amplitude of the hills in the first case, and modifying the step width and height in the second case.

We investigate how the design changes when optimized for diverse terrains with increasing difficulty levels. 
The optimization results are given in Table \ref{table:results} and Table~\ref{table:improvement}.
The general trend is that with increasing terrain roughness, the design tends to have longer legs that allow the robot to overcome obstacles in the terrain with ease.

For maximizing tracking performance ($\mathcal{C}_v$), the designs maintain the slight increase in thigh/shank ratio as in flat terrain, although augmenting the overall leg length as the terrain gets harder. This increment in limb longitude translates into a better command tracking performance as complexity levels rise, with improvements of \unit[25-27]{\%} in the extreme cases.

The results for torque and mechanical power minimization continue this trend. In addition, in these cases we clearly see the importance of $w_t$, which trades-off command tracking performance. For torque minimization, the design optimizer always seeks to reduce the link lengths as much as possible while still maintaining enough workspace to locomote through the hills and steps present in the environment. Similar behavior can be seen in the power minimization case, where the higher thigh scale ratio is preserved while increasing the leg length in the most extreme cases.

\subsection{Comparison with a Model-based Baseline}

We validate our framework by comparing with a previous work: \textit{Vitruvio}~\cite{chadwick2020Vitruvio}.
Vitruvio evaluates each design instance using a trajectory optimization method~\cite{winkler18} in the design optimization loop.
\cite{chadwick2020Vitruvio} presented leg link optimization of the ANYmal-B robot~\cite{Hutter2017ANYmal} on flat terrain with fixed forward directional command (section III-B in \cite{chadwick2020Vitruvio}).
We solve the same design optimization task using our framework, and compare the resulting designs with respect to the design objectives defined by \cite{chadwick2020Vitruvio}.

\subsubsection{Experimental setup}
We optimize the link lengths of the ANYmal-B robot for the task of forward locomotion on flat terrain. The robot is commanded to walk at \unit[0.36]{m/s} in $x$ direction.
Vitruvio introduced three different metrics to minimize: Joint torque minimization, mechanical power minimization, and \ac{MCOT} minimization. The two first metrics are the same introduced in section \ref{sec:results}-A, but without the weighting factor $w_t$. \ac{MCOT} is defined as follows:

\begin{equation*}
    MCOT =  \frac{\mathcal{P}_{mech}}{mg\lvert v \rvert}
\end{equation*}

In contrast to \cite{chadwick2020Vitruvio}. where the optimization is performed for each leg independently, we treat the system as a whole, so the \ac{MCOT} is computed using the total mass and mechanical power of the robot.

Additionally, we used neural network dynamics model of \textit{ANYdrive} actuator to enhance the simulation fidelity~\cite{Hwangbo2019RL}, such that the data used for the optimization is more realistic. Note that Vitruvio cannot take into account such a complex actuation dynamics. Our policy can be deployed on the robot while Vitruvio's motion requires additional regularization and a whole body controller~\cite{winkler18}.

\subsubsection{Results}
\begin{figure*}
    \centering
    \includegraphics[width=0.85\textwidth]{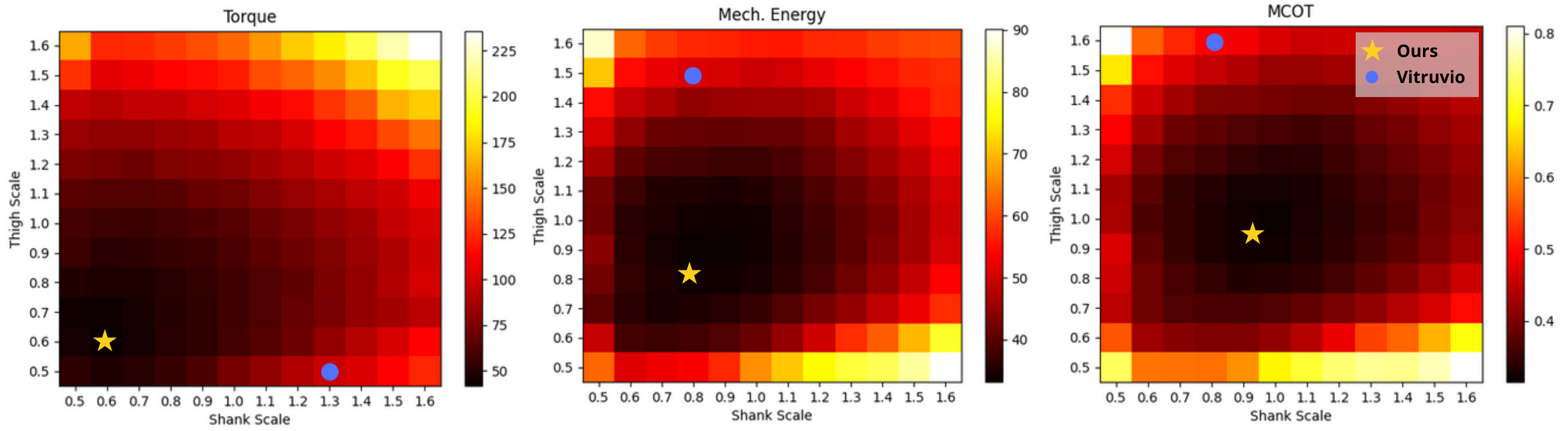}
    \caption{Cost maps of ANYmal-B for different metrics. Yellow stars represent optimal designs found with our framework, and blue dots are optimal designs reported in Vitruvio \cite{chadwick2020Vitruvio}.}
    \label{fig:anymalBmaps}
\end{figure*}

we build $12 \times 12$ cost maps over the design space for the objectives presented above (Figure \ref{fig:anymalBmaps}).
The value at each cell is the average value of 500 episodes in 500 different environments for each design instance.

As it can be seen in Fig. \ref{fig:anymalBmaps}, optimal designs by Vitruvio (blue dots) tend to have a big difference between thigh and shank lengths, and do not reside in the low-cost area of the cost map according to our simulation (dark area). In contrast, our designs (stars) do fall in these areas and present more moderate differences between thigh and shank scales.

The mismatch can come from different reasons, such as the modelling simplifications (centroidal dynamics, lack of actuator model, optimizing each leg independently) or the difference in the capability of the control methods. One important source is the kinematic restriction imposed by the pre-computed motion trajectory. Vitruvio relies on a predefined trajectory generated by trajectory optimization~\cite{winkler18}, and rejects designs that cannot fit in (e.g., too long or too short legs). 

In addition, our result is consistent with the results reported by Ha et al.~\cite{Ha2016TaskBasedDO} where it is shown by controlled experiments that the optimal design for torque minimization of two-link legs for quadrupedal robots falls within a shank/thigh ratio between 1.0 and 1.5. Results from other frameworks~\cite{dinev2021co, devincenti2021control} also follow this trend.

\subsection{Higher dimensional experiment}
\label{sec:results:gear}

To verify the effectiveness of our approach in a higher dimensional example, we also include the gear ratios for both the hip and the knee actuators as design parameters. 
This results in a 4D design space; 2 for leg lengths, 1 for knee gear ratio, and 1 for hip gear ratio. The nominal values for these parameters are 5.6 and 8.0 for the hip and knee gears, respectively.

We run the design optimization for flat terrain and rough terrains. The result is shown in Table \ref{table:results_gear}. The optimized designs have higher hip gear ratio and lower knee gear, which results in stronger hips and faster knees.

\begin{table}[t]
\centering
\caption{Optimized designs and mean improvement with respect to nominal design for 4D design space}
\label{table:results_gear}
\begin{tabular}{|c|cccccc|}
\hline
\textbf{$\bm{\mathcal{C(T)}}$}                & \textbf{Terrain} & \textbf{Thigh} & \textbf{Shank} & \textbf{\begin{tabular}[c]{@{}c@{}}Hip\\ Gear\end{tabular}} & \textbf{\begin{tabular}[c]{@{}c@{}}Knee \\ Gear\end{tabular}} & \textbf{\begin{tabular}[c]{@{}c@{}}Improve-\\ ment\end{tabular}} \\ \hline
\multirow{3}{*}{$\bm{\mathcal{C}_v}$} & Flat             & 1.06           & 1.0            & 7.05                                                        & 3.54                                                          & 0.2\%                                                            \\
                             & Mid Hills        & 1.06           & 1.04           & 7.5                                                         & 3.7                                                           & 5.84\%                                                           \\
                             & Mid Steps        & 1.07           & 1.06           & 7.7                                                         & 3.8                                                           & 10.8\%                                                           \\ \hline
\multirow{3}{*}{$\bm{\mathcal{C}_{\tau}}$} & Flat             & 0.60           & 0.65           & 9.42                                                        & 3.0                                                           & 48.8\%                                                           \\
                             & Mid Hills        & 0.60           & 0.72           & 9.85                                                        & 3.1                                                           & 42.71\%                                                          \\
                             & Mid Steps        & 0.66           & 0.83           & 9.8                                                         & 3.2                                                           & 35.02\%                                                          \\ \hline
\multirow{3}{*}{$\bm{\mathcal{C}_p}$} & Flat             & 0.75           & 0.65           & 9.0                                                         & 3.2                                                           & 18.3\%                                                           \\
                             & Mid Hills        & 0.81           & 0.62           & 9.63                                                        & 3.0                                                           & 15.88\%                                                          \\
                             & Mid Steps        & 0.94           & 0.71           & 9.8                                                         & 3.2                                                           & 9.32\%                                                           \\ \hline
\end{tabular}
\end{table}

The two additional optimized design parameters result in higher improvements in our considered optimization objectives compared to the 2D cases.

\subsection{Computational Benefit}

 Having a policy that adapts fast and performs nearly optimally for each design instance enables us to run the design optimization without training specialized policies for different designs.
 In our setup (see~\ref{setup}), Each specialized policy takes about 12 hours of wall clock time to train, while training a meta-policy takes approximately 72 hours until convergence. The design optimization runs for 30 generations, with 35 different designs per generation. This means that each generation would require about 420~h if the policy training is done in series. On the other hand, the design optimization including policy adaptation using our meta-policy takes only about 1.4~h.
\section{CONCLUSION}

We present a novel approach to the design optimization problem by introducing an adaptive \ac{RL}-based locomotion controller during the optimization process.
The locomotion policy is conditioned on the design parameters such that it can act as an optimal policy for each design instance.
We use Meta Reinforcement Learning to enable the fast adaptation of the policy to a specific design during design optimization.
The pretrained meta-policy is used for design optimization alongside a genetic algorithm and any user-defined optimization metric.
In principle, our framework can be applied to any design problem since both the controller (meta-policy) and the design optimizer (genetic algorithm) are model-free.

We would like to highlight the flexibility of our approach in considering the robot's operating environment during the design process, which can be limited in the conventional optimization-based approach, where we need analytic dynamics models.

We applied our framework to optimize leg link lengths of two quadrupedal robots. Our results show that with \ac{Meta-RL}, we can obtain a policy that achieves close-to-optimal locomotion control of the robot within a range of design specifications with only few adaptation steps. Furthermore, in contrast to model-based methods, the policy can deal with unanticipated changes in the environment.
This results in designs optimized in a more versatile sense, not overfitted to specific motions and environments. Our case studies show that a considerable improvement can be obtained compared to a hand-crafted design (nominal design).
Additionally, we conducted a qualitative comparison with an existing framework (Vitruvio) and showed that our approach results in lower-cost designs that are consistent with existing literature. 
 
One of the limitations of our framework is the cost functions used for the design optimization. Although being standard metrics in the design optimization literature, these cost functions could not capture the actual dynamics of the system. E.g., the power consumption of the physical system consists of not only the joule heating or mechanical energy, but also other factors like transmission losses that are not reflected in our cost functions. 
Furthermore, the ratio of different sources is unclear. Thus, further research in realistic cost functions is required, but this wasn't part of the scope of this project.

Future work should seek to build prototypes of optimized designs and validate them on the physical system. The addition of more design parameters, both discrete and continuous, is also a possible work direction in order to evaluate how \ac{Meta-RL} behaves with a wider parameter space. 

%%%%%%%%%%%%%%%%%%%%%%%%%%%%%%%%%%%%%%%%%%%%%%%%%%%%%%%%%%%%%%%%%%%%%%%%%%%%%%%%
%\clearpage
\section*{APPENDIX}

The reward function consists of two main terms: $r_v$ and $r_{\omega}$. These terms make the policy learn to follow the given velocity command, both linear and angular. The remaining values are regularization terms, which improve the overall quality of the motion.

\begin{table}[ht]
\centering
\caption{Reward function definition}
\label{table:reward}
\begin{tabular}[width=\columnwidt]{|cc|}
\hline
\multicolumn{2}{|c|}{\textbf{Reward Function}}                                                                                                                           \\ \hline
\multicolumn{2}{|c|}{\begin{tabular}[c]{@{}c@{}}$\mathcal{R}(s) = 0.5r_v + 0.2 r_{\omega} + 0.1 r_{v_{stability}}+ 0.1 r_{\omega_{stability}}$ \\ $+ 0.005 r_{fm} - 0.5r_{bc} - 0.05 r_{ts} - 0.005 r_{ms} - 0.001 r_{\tau}$\end{tabular}} \\ \hline
\multicolumn{1}{|c|}{\textbf{Reward term}}                                                                               & \textbf{Value}                                \\ \hline
\multicolumn{1}{|c|}{\begin{tabular}[c]{@{}c@{}}Linear velocity\\ $(r_v)$\end{tabular}}                                        & $exp\left(-1.5 \cdot \left\| v_{xy}^{target} - v_{xy} \right\|^2_2\right)$                                              \\ \hline
\multicolumn{1}{|c|}{\begin{tabular}[c]{@{}c@{}}Angular velocity\\ $(r_{\omega})$\end{tabular}}                                       &   $exp\left(-2.0 \cdot (\omega_z^{target} - \omega_z) \right)$                                           \\ \hline
\multicolumn{1}{|c|}{\begin{tabular}[c]{@{}c@{}}Linear base stability\\ $(r_{v_{stability}})$\end{tabular}}                                  & $exp\left(-1.5 \cdot (v_z)^2\right)$                                             \\ \hline
\multicolumn{1}{|c|}{\begin{tabular}[c]{@{}c@{}}Angular base stability\\ $(r_{\omega_{stability}})$\end{tabular}}                                 & $exp\left(-1.5 \cdot \left\|(\omega_{xy})\right\|_2^2\right)$                                                \\ \hline
\multicolumn{1}{|c|}{\begin{tabular}[c]{@{}c@{}}Foot motion\\ $(r_{fm})$\end{tabular}}                                            & $\sum_{i\in I_{swing}} \left(\mathbf{1}_{\mathcal{F}_{clear}}(i)/|I_{swing}|\right)$                                             \\ \hline
\multicolumn{1}{|c|}{\begin{tabular}[c]{@{}c@{}}Body collision\\ $(r_{bc})$\end{tabular}}                                         & $|I_{c,body} \setminus I_{c,foot}|$                                              \\ \hline
\multicolumn{1}{|c|}{\begin{tabular}[c]{@{}c@{}}Target smoothness\\ $(r_{ts})$\end{tabular}}                                      & $\left\|(r_{f,d})_t - 2 (r_{f,d})_{t-1} + (r_{f,d})_{t-2}\right\|$                                              \\ \hline
\multicolumn{1}{|c|}{\begin{tabular}[c]{@{}c@{}}Motion smoothness\\ $(r_{ms})$\end{tabular}}                                      & $\left\|u_t - u_{t-1}\right\|$                                              \\ \hline
\multicolumn{1}{|c|}{\begin{tabular}[c]{@{}c@{}}Torque\\ $(r_{\tau})$\end{tabular}}                                                 & $\sum_{i\in joints} |\tau_i|$                                             \\ \hline
\end{tabular}
\end{table}

\begin{table}[ht]
\centering
\caption{PPO Hyperparameters}
\label{table:ppo}
\begin{tabular}[width=\columnwidth]{|c|c|}
\hline
\textbf{Hyperparameter} & \textbf{Value} \\ \hline
Discount factor $\gamma$         & 0.993          \\
Entropy coefficient     & 0.0            \\
Adam stepsize $\alpha$           &  $5 \times 10^{-4}$         \\
GAE lambda $\lambda$              & 0.95           \\
Clipping parameter      & 0.2            \\
Meta-batch size M & 5 \\
Num. Mini-batches      & 10             \\ \hline
\end{tabular}
\end{table}

%%%%%%%%%%%%%%%%%%%%%%%%%%%%%%%%%%%%%%%%%%%%%%%%%%%%%%%%%%%%%%%%%%%%%%%%%%%%%%%%

%\section*{ACKNOWLEDGMENT}

%%%%%%%%%%%%%%%%%%%%%%%%%%%%%%%%%%%%% REFERENCES %%%%%%%%%%%%%%%%%%%%%%%%%%%%%%%

% Command included with the template

\bibliographystyle{bibliography/IEEEtran}
\bibliography{bibliography/references}

\end{document}